\documentclass[10pt,conference,a4paper]{IEEEtran}
\usepackage{graphicx}
\usepackage{multirow}
\usepackage{amsmath}
\usepackage{bm}
\usepackage{cite}
\usepackage{url}
\usepackage{algorithm}
\usepackage{algorithmic}
\usepackage{subfig}
\usepackage{mathrsfs}
\begin{document}
\title{Fully Convolutional Recurrent Network for Handwritten Chinese Text Recognition}

\author{\IEEEauthorblockN{Zecheng Xie, Zenghui Sun, Lianwen Jin$^\ast$, Ziyong Feng, Shuye Zhang}
\IEEEauthorblockA{
College of Electronic and Information Engineering\\
South China University of Technology\\
Guangzhou, China\\
xiezcheng@foxmail.com, $^\ast$lianwen.jin@gmail.com, sunfreding@gmail.com, \\f.ziyong@mail.scut.edu.cn, shuye.cheung@gmail.com }
}

\maketitle

\begin{abstract}
This paper proposes an end-to-end framework, namely fully convolutional recurrent network (FCRN) for handwritten Chinese text recognition (HCTR).
Unlike traditional methods that rely heavily on segmentation, our FCRN is trained with online text data directly and learns to associate the pen-tip trajectory with a sequence of characters.
FCRN consists of four parts: a path-signature layer to extract signature features from the input pen-tip trajectory, a fully convolutional network to learn informative representation, a sequence modeling layer to make per-frame predictions on the input sequence and a transcription layer to translate the predictions into a label sequence.
%The FCRN is end-to-end trainable in contrast to conventional methods whose components are separately trained and tuned.
We also present a refined beam search method that efficiently integrates the language model to decode the FCRN and significantly improve the recognition results.

We evaluate the performance of the proposed method on the test sets from the databases CASIA-OLHWDB and ICDAR 2013 Chinese handwriting recognition competition, and both achieve state-of-the-art performance with correct rates of 96.40\% and 95.00\%, respectively.
\end{abstract}


% \fi
\IEEEpeerreviewmaketitle

\section{Introduction}
Handwritten Chinese text recognition (HCTR) is a challenging problem and has received intensive concerns from numerous researchers.
The large character set, diversity of writing styles and character-touching problem are the main difficulties of HCTR.
Traditional methods\cite{zhou2013handwritten}\cite{zhou2014minimum} overcome these difficulties by integrating segmentation and recognition.
Generally, a segmentation-recognition candidate lattice\cite{zhou2013handwritten} is first derived from the input pen-tip trajectory through operations of over-segmentation, component combination and character recognition.
Based on the lattice, the optimal path can be searched by simultaneously considering the character recognition score, in addition to the geometric and linguistic contexts.
Zhou et al.\cite{zhou2013handwritten} proposed a method based on semi-Markov conditional random fields, which combined candidate character recognition scores with geometric and linguistic contexts.
Zhou et al.\cite{zhou2014minimum} described an alternative parameter learning method, which aimed at minimizing the character error rate rather than the string error rate.
Vision Objects Ltd., France, whose system yielded the best performance in the ICDAR 2013 Chinese handwriting recognition competition\cite{yin2013icdar}, introduced three `experts' that were responsible for segmentation, recognition and interpretation.
They employed a global discriminant training scheme on the text level to learn the classifier parameter and meta-parameters of the recognizer.

However, traditional methods based on over-segmentation can barely overcome their own limitations to rectify the mis-segmentations when characters are not correctly separated.
Segmentation-free models\cite{liwicki2007novel,graves2009novel,messina2015segmentation,shi2015end,he2015reading}
%\cite{liwicki2007novel}\cite{graves2009novel}\cite{messina2015segmentation}\cite{shi2015end}\cite{he2015reading}
have been studied and have proved to be useful in different areas.
Liwicki et al.\cite{liwicki2007novel} and Graves et al.\cite{graves2009novel} combined bidirectional long short-term memory (LSTM) and the connectionist temporal classifier (CTC) to build a speech recognizer.
%\cite{liwicki2007novel}\cite{graves2009novel} combined bidirectional long short-term memory (LSTM) and connectionist temporal classifier (CTC) to build a speech recognizer.
Messina et al.\cite{messina2015segmentation} applied multi-dimensional LSTM with CTC to offline HCTR.
Recently, Shi et al. [7] proposed a network architecture called convolutional recurrent neural network (CRNN), which consists of the convolutional layers, recurrent layers and transcription layer, for image-based sequence recognition.

Similar to the aforementioned tasks, variable-length input is also the fundamental difficulty when solving HCTR problems.
In this paper, we propose a fully convolutional recurrent network (FCRN), which is a novel framework for HCTR problems that possesses the following advantages:
%distinctive advantages over conventional approaches:
(1) It applies a path-signature layer to generate signature feature maps for online data, which uniquely characterizes the pen-tip trajectory.
(2) It takes an input sequence of arbitrary length and outputs a corresponding label sequence without pre-segmentation.
%requiring any prior segmentation.
(3) It is end-to-end trainable. All its components can be jointly trained to fit each other and improve the overall function and reliability.
Language models are of great importance for speech recognition and online text recognition, and have been proved to be effective by Wang et al.\cite{wang2012handwritten} and Wu et al.\cite{wu2015evaluation}.
In this paper, we adopted a refined beam search method to integrate a language model to decode our FCRN.
Experiments showed that by incorporating lexical constraints and prior knowledge about a certain language, the language model can further decrease the error rate by 2\%-5\%.
%improves system performance.
\begin{figure*}[th]
\centering
\includegraphics[width=\textwidth]{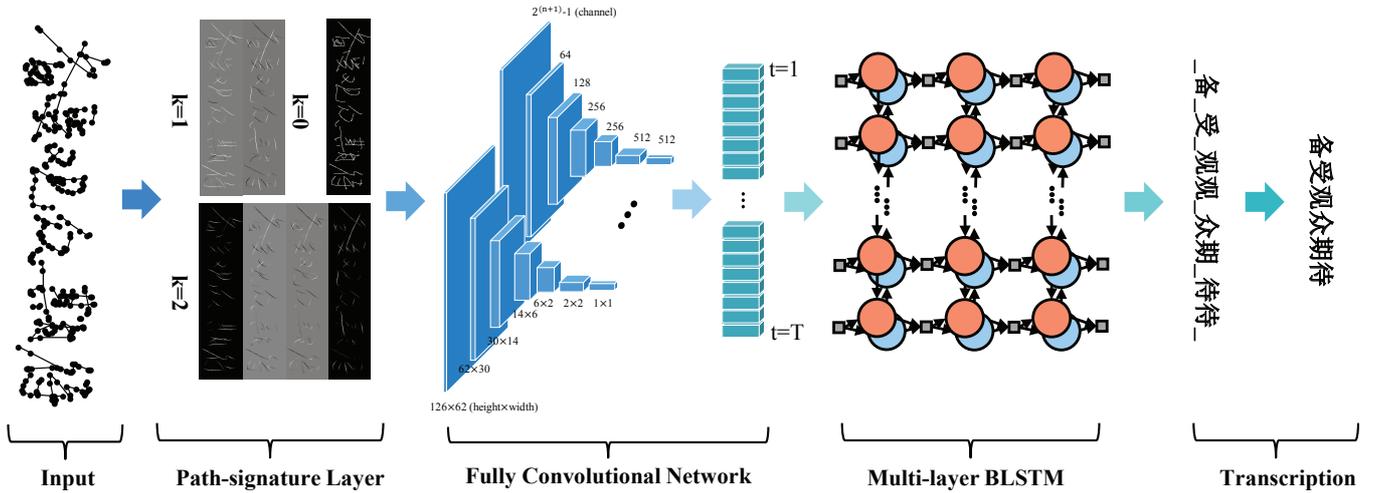}
\caption{Architecture of the proposed fully convolutional recurrent network. Given the input pet-tip trajectory, path-signature layer extracts $2^{n+1}-1$ signature feature maps with informative dynamics. Then a fully convolutional network produces a length T feature sequence whose frames correspond to receptive fields with height 126 pixels and width 62 pixels on the signature feature maps.
After that, a multi-layer BLSTM predicts a probability distribution for each frame in the feature sequence. Finally, transcription layer derives a label sequence from the per-frame predictions.}
\label{FigureOverallSystem}
\end{figure*}

The remaining parts of the paper are organized as follows. In Section 2, we illustrate the framework of FCRN in detail. In Section 3, we describe language modeling and in Section 4, we present the experimental results. In Section 5, we conclude the paper.

%\begin{figure}[htbp]
%\centering
%\includegraphics{input\frame_work.eps}
%\caption{frame_work}
%\label{fig:test}
%\end{figure}

\section{Fully Convolutional Recurrent Network}
Given the training set $Q$ and a training instance $(x,z)$ in which $x$ represents the pen-tip trajectory and $z$ is the corresponding label sequence, the FCRN aims to minimize the loss function $L(Q)$ as the negative log probability of correctly labelling all the training examples in $Q$:
%Given the training set $Q$ and $(x,z)$ represents the pair of pen-tip trajectory and corresponding label sequence, the FCRN aims to minimize the loss function $L(Q)$ as the negative log probability of correctly labelling all the training examples in $Q$:
\begin{equation}
    L(Q)=-\ln \prod_{(x,z)\subset Q} p(z|x) = - \sum_{(x,z)\subset Q} \ln p(z|x).
\end{equation}

Fig.~\ref{FigureOverallSystem} describes the network architecture of the proposed FCRN.
The FCRN consists of four components.
First, the path signature layer outputs $2^{n+1}-1$ feature maps that are used to characterize the pen-tip trajectory from the online handwritten text data.
Second, a fully convolutional network (FCN) produces a feature sequence in which each frame represents the feature vector of a $126 \times 62$ receptive field on the signature feature maps.
Third, multi-layer bidirectional LSTM (BLSTM) predicts a probability distribution for each frame in the feature sequence.
Finally, the transcription layer derives a label sequence from the per-frame predictions.

\subsection{Path signature layer}
%Path-signature, pioneered by K.T. Chen\cite{chen1958integration} in the form of iterated integrals, can extracts sufficient information that uniquely characterizes paths (e.g., in online handwriting) of finite length. Path-signature was first introduced into handwritten Chinese character recognition by Benjamin Graham\cite{graham2013sparse}, and improved the accuracy of online character recognition.
The path signature, pioneered by Chen\cite{chen1958integration} in the form of iterated integrals and developed by Terry Lyons and his colleagues to play a fundamental role in rough theory\cite{hambly2010uniqueness,lyonssystem,lyons2014rough}, can extract sufficient information that uniquely characterizes paths (e.g., in online handwriting) of finite length.

Assume a time interval $[T_1,T_2]$ and the writing plane $U \subset R^2$. Then a pen stroke can be expressed as: $S:[T_1,T_2]\rightarrow U$. For intervals $[t_1,t_2]\subset [T_1,T_2]$, the $k$-th iterated integral of $S$ is the $2^k$ dimensional vector defined by
\begin{equation}
    S_{t_1,t_2}^k = \int_{t_1<q_1<\cdots <q_k<t_2} 1dS_{q_1} \otimes,\cdots,\otimes dS_{q_k}.
\end{equation}
By convention, the $k$ = 0 iterated integral is simply the number one (i.e., the offline map of the character), the $k$ = 1 iterated integral represents the path displacement, and the $k$ = 2 iterated integral represents the curvature of the path.

%Note that the iterated integrals increase rapidly in dimension as $k$ increases while carrying very little information.
Note that the $k$-th iterated integral of $S$ increases rapidly in dimension as $k$ increases while carrying very little information.
Hence, a truncated signature is preferred. If truncated at level $n$, the path signature can be expressed by
\begin{equation}
    P(S)_{t_1,t_2}^n = (1,S_{t_1,t_2}^1,\cdots ,S_{t_1,t_2}^n).
\end{equation}
The dimension of the truncated path signature is $2^{(n+1)}-1$ (i.e., the number of feature maps). When $S$ is a straight line, the iterated integrals $S_{t_1,t_2}^k$ can be calculated using
\begin{equation}
    \begin{split}
        S_{t_1,t_2}^0 &= 1, \\
        S_{t_1,t_2}^1 &=\bigtriangleup_{t_1,t_2}, \\
        S_{t_1,t_2}^2 &=(\bigtriangleup_{t_1,t_2} \otimes \bigtriangleup_{t_1,t_2})/2!,\\
        S_{t_1,t_2}^3 &=(\bigtriangleup_{t_1,t_2} \otimes \bigtriangleup_{t_1,t_2} \otimes \bigtriangleup_{t_1,t_2} )/3!,\cdots,
    \end{split}
\end{equation}
%\begin{gather}
%    %\begin{split}
%        S_{t_1,t_2}^0 = 1, S_{t_1,t_2}^1=\bigtriangleup_{t_1,t_2}, \notag\\
%        S_{t_1,t_2}^2=(\bigtriangleup_{t_1,t_2} \otimes \bigtriangleup_{t_1,t_2})/2!,\\
%        S_{t_1,t_2}^3=(\bigtriangleup_{t_1,t_2} \otimes \bigtriangleup_{t_1,t_2} \otimes \bigtriangleup_{t_1,t_2} )/3!,\cdots, \notag
%    %\end{split}
%\end{gather}
where $\bigtriangleup_{t_1,t_2}:=S_{t_2}-S_{t_1}$ denotes the path displacement.
Fig.~\ref{FigureOverallSystem} shows the signature feature maps of online text data to better illustrate the idea of the path signature.

\subsection{Fully convolutional network}
A convolutional network is a powerful visual model that extracts high-level abstract features from an image.
Inheriting this property, an FCN\cite{long2015fully} takes an input image of arbitrary size and outputs a corresponding-sized dense response map.
Unlike image cropping or sliding window-based approaches, an FCN eliminates redundant computations by sharing a convolutional response map layer-by-layer to make inference and backpropagation efficient.

Basic operations in a convolutional network, such as convolution, pooling and the element-wise activation function, are translation invariant.
Therefore, locations in the last response map correspond to rectangular regions that are called the receptive field in the original image to which they are associated.
Layer-wise formulations to calculate the exact location and size of the receptive field are provided below:
\begin{gather}
    r_i=(r_{i+1}-1)\times s_i+k_i,\\
    p_i=s_i \times p_{i+1}+(\frac{k_i-1}{2}-d_i),
\end{gather}
where $r_i$ is the local region size of the $i$-th layer, $k$ is the kernel size, $s$ is the stride size, $p$ denotes the position and $d$ is the padding size of a particular layer.

\begin{figure}[t]
\centering
\includegraphics[width=0.4\textwidth]{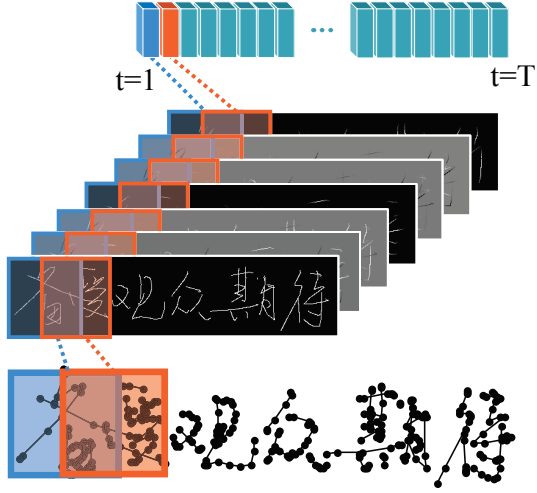}
\caption{The receptive field. Successive frames in the output feature
sequence of FCN correspond to the overlapped receptive fields on the
original data.}
\label{FigureTheReceptiveField}
\end{figure}

The FCN takes the input of the signature feature maps and outputs a length T feature sequence. As shown in Fig.~\ref{FigureTheReceptiveField}, successive frames in the output feature sequence correspond to the overlapped receptive fields on the original data.

\subsection{Multi-layer BLSTM}
\begin{figure}[t]
\centering
\includegraphics[width=0.4\textwidth]{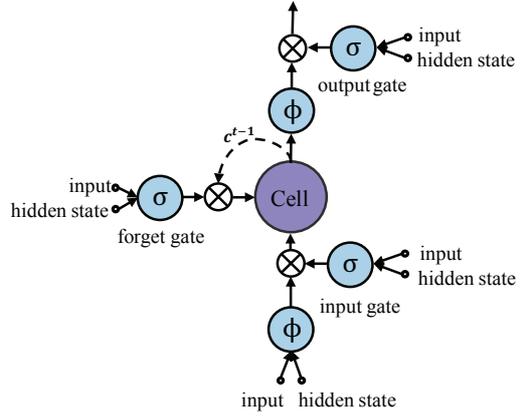}
\caption{Long short-term memory (LSTM) cell.}
\label{LSTM}
\end{figure}

The traditional recurrent neural network (RNN) is well known for its self-connected hidden layer that recurrently transfers information from output to input.
However, the traditional RNN suffers from gradient vanishing and exploding problem.
Long Short-Term Memory (LSTM)\cite{hochreiter1997long}, the core of which is the memory cell and three gates (as illustrated in Fig.~\ref{LSTM}), is used here for its strong ability to capture complex and long-term temporal dynamics.
In particular, the three sigmoidal nonlinear gates, namely the input gate, forget gate and output gate, control the information flow in and out of the cell unit.
The input gate protects the cell unit from the influence of the current input along with past hidden states,
the forget gate allows the memory cell to forget or maintain its previous states
and the output gate decides how much memory is to be sent out as hidden states.

To capture complex long-term dependencies, we adopted LSTM for modeling the input feature sequence produced by FCN.
Each time it receives a frame from the input feature sequence, LSTM updates its hidden states and predicts a distribution for further transcription. We note that LSTM has the following properties for the HCTR problem.
Shi et al.\cite{shi2015end} showed that LSTM naturally captures the contextual information from a sequence, which makes the text recognition process more efficient and reliable than processing each character independently.
Moreover, LSTM is not limited to fixed length inputs or outputs, which allows for modeling sequential data of arbitrary length.
%Furthermore, with an FCN, LSTM is straightforward to be trained end-to-end.
Furthermore, LSTM can be jointly trained with an FCN in a unified network (e.g., FCRN).
Joint training can benefit both the convolutional layers and LSTM, and improve overall text recognition performance. %while eliminating the need for complex multi-step pipeline in HCCR.

% In this paper, we use the vector formulas for the LSTM unit update suggested by \cite{zaremba2014learning}:
% \begin{align}
%     i_t &= \sigma(I_{xi}x_t+H_{hi}h_{t-1}+b_i)\\
%     o_t &= \sigma(I_{xo}x_t+H_{ho}h_{t-1}+b_o)\\
%     f_t &= \sigma(I_{xf}x_t+H_{hf}h_{t-1}+b_f)\\
%     g_t &= \sigma(I_{xg}x_t+H_{hg}h_{t-1}+b_g)\\
%     c_t &= f_t \odot c_{t-1} + i_t \odot g_t  \\
%     h_t &= o_t \odot \sigma(c_t)
% \end{align}

Standard LSTM can only use past contextual information in one direction. This is far from sufficient for HCTR in which bidirectional contextual knowledge is accessible.
Bidirectional LSTM (BLSTM) can learn long-range context dynamics in both input directions and significantly outperform unidirectional networks.
Furthermore, as suggested by Pascanu et al.\cite{pascanu2013construct}, we stack multiple BLSTMs in our framework to capture higher-level abstract information for further transcription. 
%In this paper, we adopted a three-layered BLSTM, which is illustrated in Fig.~\ref{FigureMultiLayerBLSTM}.
Finally, fully connected layers were incorporated between the BLSTM and transcription layer to enhance classification.

%\begin{figure}[t]
%\centering
%\includegraphics[width=0.\textwidth,angle=90]{test21.eps}
%\caption{frame work}
%\label{fig:test}
%\end{figure}

%
%\begin{figure}[t]
%\centering
%\subfloat[]{
%\label{FigureCell}
%%\includegraphics[angle=90,width=0.25\textwidth]{test42.eps}
%\includegraphics[width=0.25\textwidth]{test53.eps}
%}
%%\hspace{80pt}
%\subfloat[]{
%\label{FigureMultiLayerBLSTM}
%\includegraphics[width=0.2\textwidth]{test38.eps}
%}
%\caption{(a) Long short-term memory (LSTM) cell. (b) Multi-layer bidirectional LSTM.}
%\end{figure}
%
%\begin{figure}[t]
%\centering
%\begin{minipage}[t]{0.25\textwidth}
%\centering
%\includegraphics[angle=90,width=\textwidth]{test21.eps}
%\caption{Long Short-term Memory cell}
%\end{minipage}
%\begin{minipage}[t]{0.2\textwidth}
%\centering
%\includegraphics{test22.eps}
%\caption{Multi-layer Bidirectional LSTM}
%\end{minipage}
%\end{figure}

\subsection{Transcription}
Traditional approaches for HCTR are confronted with the paradox of a circular dependency between segmentation and recognition.
To avoid the difficulty of segmentation, we adopted connectionist temporal classification (CTC) as the transcription layer in our framework. CTC allows an FCN and LSTM for sequential training without requiring any prior alignment between input images and their corresponding label sequences.
%Connectionist Temporal Classification (CTC), which allows FCN and LSTM for sequential training without requiring any prior alignment between input images and their corresponding label sequences, is adopted as the transcription layer in our framework.

We denote the character set as $C'=C \cup \{blank\}$, where $C$ contains all characters used in this task and `blank' represents the null emission.
Given length $T$ input sequences $\bm{s}=s_1,s_2,\cdots,s_T$, where $s_t \in R^{|C'|}$, we can obtain an exponentially large number of length $T$ label sequences, known as alignments, by assigning each time step a label and concatenating the labels to form a label sequence.
The alignments are denoted by $\bm{\pi}$ and their probability is given below:
\begin{equation}
    Pr(\bm{\pi}|\bm{s}) = \prod_{t=1}^T Pr(\pi_t,t|\bm{s}).
\end{equation}
By applying a sequence-to-sequence operation $\mathscr{B}$, alignments can be mapped onto a transcription (denoted by $\bm{l}$) by first removing the repeated labels and then the blanks. For example, `apple' can be transformed by $\mathscr{B}$ from `\_aa\_p\_pl\_ll\_e' or `\_a\_pp\_p\_l\_ee\_'. The total probability of a transcription can be calculated by summing the probabilities of all alignments that correspond to it:
\begin{equation}
\label{EquationAligment}
    Pr(\bm{l}|\bm{s}) = \sum_{\bm{\pi}:\mathscr{B}(\bm{\pi})=\bm{l}} Pr(\bm{\pi}|\bm{s}).
\end{equation}
As described by Graves and Jaitly\cite{graves2014towards}, because we do not know the exact position of the labels within a particular transcription, we consider all locations where they could occur;
that is, what allows a CTC to train a network without pre-segmented data.
A detailed forward-backward algorithm to efficiently calculate the probability in Eq.~\eqref{EquationAligment} was described by Graves\cite{graves2012supervised}.

\section{Language Modeling}
The statistical language model plays a significant role in many technological applications, including online and offline handwritten text recognition, speech recognition and language translation.
The statistical model of language (e.g., a length T sequence of words) is represented as follows:
\begin{equation}
    \label{EquationLanguageModel}
    Pr(w_{1}^T)=\prod_{t=1}^T Pr(w_t|w_{1}^{t-1}),
\end{equation}
where $w_t$ is the $t$-th word in the sequence and $w_{i}^j$ denotes the sequence $(w_i,w_{i+1},\cdots,w_{j-1},w_{j})$.
%In practice, n-gram model that considers the conditional probability of the next word given the last $n-1$ words is more often adopted for language modeling:
In fact, closer words in a word sequence tend to be more dependent.
Therefore, n-gram model, which is constructed by the conditional probability of the next word given the last $n-1$ words, is more often used in practice:
%Therefore, the conditional probability of the next word given the last $n-1$ words, which constructs the n-gram model, is more often used in practice:
%In practice, we usually use n-gram model for language modeling. It considers the conditional probability of the next word given the last $n-1$ words.
%In fact, closer words in a words sequence tends to be more dependent
%truncated version of conditional probability is consider in practice.
\begin{equation}
    Pr(w_t|w_{1}^{t-1}) \approx Pr(w_t|w_{t-n+1}^{t-1}).
\end{equation}
In this paper, we only considered the character bigram and trigram language model in the experiments.

Decoding a CTC network can be easily accomplished through `naive decoding'\cite{graves2012supervised}, which takes labels within the highest probability for each frame and obtains the transcription by applying operation $\mathscr{B}$ to the alignment.
However, naive decoding is not sufficient and can be improved by language modeling.
By incorporating lexical constraints and prior knowledge about the language, language modeling can rectify some obvious semantic errors, and thus improves the recognition result.
To integrate the language model and overcome the difficulty that the operation $\mathscr{B}$ creates, we adopted a refined beam search method to decode the FCRN.
Specifically, given the per-frame prediction distribution from the Multi-layer BLSTM,
the beam search method first selects the candidates with confidence scores higher than a probability threshold for each time step, and then in the next steps it determines the time steps with one and only one candidate `blank' to separate the alignments into regions.
%Then it finds out the time steps with one and only one candidate 'blank' to separate the alignments into regions.
Finally, it enumerates the candidate paths in every region and sequentially concatenates the candidate paths from the first to last region.
In each concatenating step, both the confidence score and the language model score of the paths is considered and only the top N paths remain for the subsequent concatenating steps.
%%$W$ denoted the threshold,we eliminate the candidate label that has probability lower than $W$.

\section{Experimental Results and Analysis}
\subsection{Online handwritten text data}
CASIA-OLHWDB\cite{liu2011casia} is a Chinese handwriting database that is often used for online Chinese handwriting recognition.
It contains both isolated characters and unconstrained text lines.
The training set of CASIA-OLHWDB for online handwritten text recognition contains 4072 pages of handwritten texts, which incorporates 41,710 text lines, including 1,082,220 characters of 2650 classes,
whereas the test set (denoted as D-Casia) contains 1020 text pages, including 269,674 characters of 2631 classes.
We randomly split the training set into two groups, with approximately 90\% for training and the remainder for gauging the convergence of the training process and further parameter learning for the beam search.
Furthermore, we assessed our proposed method on the test data (denoted as D-Com) of the online handwritten text recognition task of the ICDAR 2013 Chinese handwriting recognition competition\cite{liuicdar}, which contains 3432 text lines, including 91,576 characters of 1375 classes.
However, our actual evaluation dataset is smaller than the reported one because we removed outlier characters that are never seen in the training data, and actually contains 89,723 characters of 1258 classes.
%type	Detaied configurations	Stacked times
%Transcription	Label sequence	
%IP	n: 2048	*2
%BLSTM	c: 1024	*3
%convolution	k:2*2, s:1*1, p: 0*0	*1
%convolution	k:3*1, s:3*1, p: 0*0	*1
%pooling	k:2*2, s:2*2	*4
%
%convolution	k:3*3, s:1*1, p: 1*0	
%Signature	128*576(train) 128*2400(test)	
%input	Pen-tip trajectory	

\subsection{Textual data}
\begin{table}[b]
\caption{Character information in the corpora}
\label{TableCorpora}
\centering
\begin{tabular}{c|cc}
\hline
corpora&\#characters&\#class\\
\hline
PTR&2,199,492 &4,689\\
PH&3,697,028&4,722\\
SLD&56,279,692&6,882\\
\hline
\end{tabular}
\end{table}
The experiments were conducted on three corpora,
including the PFR corpus\cite{PFR}, which is news text from the 1998 People' s Daily corpus;
the PH\cite{PH} corpus, which is news text from the People's Republic of China's Xinhua news written between January 1990 and March 1991;
and the SLD corpus\cite{SG}, which contains news text from 2006 Sogou Lab Data.
Because the total amount of Sogou Lab Data was too large, we only used an extract in our experiments.
Detailed information about these corpora is illustrated in Table~\ref{TableCorpora}.

We constructed our language models using the SRILM toolkit\cite{stolcke2002srilm}. We built three language models based on these three corpora, and compared their roles in decoding the FCRN with the beam search method.

\subsection{Experimental setting}
\begin{table}[t]
\caption{Detailed settings of our system}
\label{TableFCRNArchitecture}
\begin{tabular}{|c|c|c|}
\hline
Layer type            &Settings               &Stack times    \\
\hline
transcription   &sequence labeling          &$\times 1$     \\
\hline
inner product   &n: 2048                     &$\times 2$     \\
\hline
BLSTM           &c: 1024                     &$\times 3$     \\
\hline
convolution     &k: $2\times 2$, s: $1\times 1$, p: $0\times 0$     &$\times 1$  \\
\hline
convolution     &k: $3\times 1$, s: $3\times 1$, p: $0\times 0$     &$\times 1$  \\
\hline
pooling         &k: $2\times 2$, s: $2\times 2$                   &\multirow{2}{*}{$\times 4$}  \\
\cline{1-2}
convolution     &k: $3\times 3$, s: $1\times 1$, p: $0\times 1$      &          \\
\hline
path-signature  &$128\times 576$ (train), $128\times 2400$ (test)  &$\times 1$ \\
\hline
input           &pen-tip trajectory                             &$\times 1$  \\
\hline
\end{tabular}
\end{table}
The detailed architecture of our FCRN for HCTR is listed in Table~\ref{TableFCRNArchitecture}.
The kernel number of each layer in our FCN from bottom to top is 64, 128, 256, 256, 512 and 512.
We also applied batch normalization\cite{ioffe2015batch} to the last four convolutional layers to enable them to converge faster and avoid over-fitting.
%Our FCRN uses path-signature to derive feature maps from pen-tip trajectory.
To accelerate the training process, we trained our network with shorter texts segmented from text lines in the training data, which could be normalized to the same height of 128 pixels, while retaining the width at fewer than 576 pixels.
In the test phase, we maintained the same height but increased the width to 2400 pixels to contain the text lines from the test set.

We constructed our FCRN network within the CAFFE\cite{jia2014caffe} deep learning framework, in which LSTM is implemented by Venugopalan et al.\cite{venugopalan2015sequence} and others are contributed by ourselves.
The optimization algorithm was AdaDelta with $\rho$=0.9. We trained our FCRN with GeForce Titan-X GPUs and it took approximately four days to reach convergence.

We used the correct rate(CR) and accuracy rate(AR) performance measurement discussed in the ICDAR 2013 Chinese handwriting recognition competition\cite{liuicdar} to assess our framework.
%Let $N$ denote the number of characters in the line while $S$, $D$, and $I$ denote the number of substitutions, deletions and insertions, respectively, then CR and AR can be defined as follows:
%\begin{gather}
%    CR = \frac{N-S-D}{N},\\
%    AR = \frac{N-S-D-I}{N}.
%\end{gather}

\subsection{Experimental results}
We compared the path signatures (Sig0, Sig1, Sig2, and Sig3) in different truncated versions on our network.
Table~\ref{TableSigResult} presents the results of our system with naive decoding (i.e., without language modeling).
We observed that Sig2 outperformed the other signatures for both CR and AR, which suggests that Sig2 already extracts sufficient information for characterizing the pen-tip trajectory.
Moreover, as the path signature increase from Sig0 to Sig2, system performance improved monotonically from 90.94\% to 94.52\% because the path signature captured better informative features from the pen-tip trajectory with higher iterated integrals.
%However,although Sig3 can bring slight more information than Sig2 theoretically, it reduce the importance the informative feature map from Sig2 by introducing 8 more feature maps, which explain why Sig3 perform worse than Sig2.
%However, although Sig3 can bring slightly more information than Sig2 theoretically, it introduces much more feature maps and reduces the weights of the informative feature maps from Sig2, which explain why Sig3 performs worse than Sig2.
%However, Sig3 performs worse than Sig2 in the experiments, therefore Sig2 is adopted in our FCRN for the following experiments.
However, Sig3 performs worse than Sig2 in the experiment, because Sig3 captures slightly more information than Sig2 but may bring much more useless feature.
%therefore Sig2 is adopted in our FCRN for the following experiments.
%Because Sig2 performed the best in all iterated integrals, we adopted it in our FCRN for the following experiments.
%This is because Sig3 requires a larger FCN to model, but we adopted a small FCN network in our experiment to balance between GPU memory consumption and system performance.
Experiments also showed that FCRN performed much better on dataset D-Casia than D-Com because the per-character sample distribution of the training set was more similar to dataset D-Casia than D-Com.

\begin{table}[t]
\caption{Correct rate and accuracy rate (\%) on dataset D-Casia and D-Com with the path signatures in different truncated versions (without language modeling).}
\label{TableSigResult}
\centering
\begin{tabular}{c|c|cc|cc}
\hline
\multirow{2}{*}{Path signatures}&\multirow{2}{*}{Feature maps}&\multicolumn{2}{c|}{D-Casia}&\multicolumn{2}{c}{D-Com}\\
\cline{3-6}
&&CR&AR&CR&AR\\
\hline
Sig0&1& 90.94& 89.86&84.91&83.55\\
Sig1&3& 93.56&93.04 &87.05&86.32\\
Sig2&7& \textbf{94.52}&\textbf{93.22} &\textbf{89.86} &\textbf{88.28}\\
Sig3&15& 93.92&93.02 &88.46&87.36\\
\hline
\end{tabular}
\end{table}
%\begin{table}[t]
%\caption{Correct Rate and Accuracy Rate(\%) on database D-Casia and D-Com based on Sig2 baseline model which integrate language model with different corpus}
%\label{TableCoporaResult}
%\centering
%\begin{tabular}{cc|cc|cc}
%\hline
%\multirow{2}{*}{Corpora}&\multirow{2}{*}{n-gram order}&\multicolumn{2}{c|}{D-Casia}&\multicolumn{2}{c}{D-Com}\\
%\cline{3-6}
%&&CR&AR&CR&AR\\
%
%\hline
%\multicolumn{2}{c|}{baseline}&94.52&93.22&89.86&88.28\\
%\hline
%\multirow{2}{*}{PTR}&2&95.93&94.62&92.97&91.36\\
%&3&96.20&94.05&93.41&90.68\\
%\hline
%\multirow{2}{*}{PH}&2&95.93&94.62&92.97&91.36\\
%&3&96.30&94.05&93.39&90.71\\
%\hline
%\multirow{2}{*}{SLD}&2&96.21&95.06&94.23&93.19\\
%&3&96.58&94.74&95.00&92.88\\
%
%%PH&96.30&94.05&93.39&90.71\\
%%PH&96.58&94.74&95.00&92.88\\
%\hline
%\end{tabular}
%\end{table}
%\begin{table}[t]
%\caption{Correct Rate and Accuracy Rate(\%) on database D-Casia and D-Com based on Sig2 baseline model which integrate language model with different corpus}
%\label{TableCoporaResult}
%\centering
%\begin{tabular}{cc|cc|cc}
%\hline
%\multirow{2}{*}{Corpora}&\multirow{2}{*}{n-gram order}&\multicolumn{2}{c|}{D-Casia}&\multicolumn{2}{c}{D-Com}\\
%\cline{3-6}
%&&CR&AR&CR&AR\\
%
%\hline
%\multicolumn{2}{c|}{baseline}&94.52&93.22&89.86&88.28\\
%\hline
%\multirow{2}{*}{PTR}&2&95.93&94.62&92.97&91.36\\
%&3&96.15&94.93&93.10&91.55\\
%\hline
%\multirow{2}{*}{PH}&2&95.97&94.67&92.76&91.17\\
%&3&96.15&94.93&93.10&91.55\\
%\hline
%\multirow{2}{*}{SLD}&2&96.21&95.06&93.41&92.01\\
%&3&96.67&95.61&94.23&93.19\\
%
%%PH&96.30&94.05&93.39&90.71\\
%%PH&96.58&94.74&95.00&92.88\\
%\hline
%\end{tabular}
%\end{table}

\begin{table}[t]
\caption{Correct rate and accuracy rate (\%) on dataset D-Casia and D-Com based on FCRN with Sig2 which integrates the language model with different corpus}
\label{TableCoporaResult}
\centering
\begin{tabular}{cc|cc|cc}
\hline
\multirow{2}{*}{Corpora}&\multirow{2}{*}{n-gram order}&\multicolumn{2}{c|}{D-Casia}&\multicolumn{2}{c}{D-Com}\\
\cline{3-6}
&&CR&AR&CR&AR\\

\hline
\multicolumn{2}{c|}{FCRN}&94.52&93.22&89.86&88.28\\
\hline
\multirow{2}{*}{PTR}&2&95.66&94.35&92.97&91.36\\
&3&95.88&94.66&93.10&91.55\\
\hline
\multirow{2}{*}{PH}&2&95.70&94.40&92.76&91.17\\
&3&95.88&94.66&93.10&91.55\\
\hline
\multirow{2}{*}{SLD}&2&95.94&94.79&93.41&92.01\\
&3&\textbf{96.40}&\textbf{95.34}&\textbf{95.00}&\textbf{92.88}\\
%&3&96.40&95.34&94.67&93.37\\

%PH&96.30&94.05&93.39&90.71\\
%PH&96.58&94.74&95.00&92.88\\
\hline
\end{tabular}
\end{table}

\begin{table}[b]
\caption{Comparison with state-of-the-art methods based on correct rate and accuracy rate (\%) on dataset D-Casia and D-Com}
\label{TableFinalResult}
\centering
\begin{tabular}{c|c|cc}
\hline
Dataset&Methods&CR&AR\\
\hline
\multirow{5}{*}{D-Casia}&Zhou et al., 2013\cite{zhou2013handwritten}&94.34&93.75 \\
&Zhou et al., 2014\cite{zhou2014minimum}&95.32 & 94.69\\
&CRNN\cite{shi2015end}& 90.94& 89.86\\
&FCRN& 94.52& 93.22\\
&FCRN with SLD corpus& \textbf{96.40}&\textbf{95.34} \\
\hline
\multirow{6}{*}{D-Com}&Zhou et al., 2013\cite{zhou2013handwritten}&94.62&94.06\\
&Zhou et al., 2014\cite{zhou2014minimum}&94.76&94.22\\
&VO-3\cite{yin2013icdar}&\textbf{95.03}&\textbf{94.49} \\
&CRNN\cite{shi2015end}& 84.91&83.55\\
&FCRN& 89.86&88.28\\
&FCRN with SLD corpus&95.00&92.88\\
%&FCRN with SLD corpus&94.67&93.37\\
\hline
\end{tabular}
\end{table}

Because Sig2 performed the best in all iterated integrals, we adopted it in our FCRN for the following experiments.
We investigated the performance of decoding FCRN with different language models using the beam search method. Table~\ref{TableCoporaResult} shows that by integrating the corpus PTR with the bigram language model, the CRs on dataset D-Casia and D-Com are increased by 1.14\% and 3.11\%, respectively and the ARs increased by 1.13\% and 3.08\%, respectively, proving the effectiveness of the language model. Experiments also showed that with a higher-order language model (e.g., trigram), our system still improved performance. %The corpus PH have a similar effect in decoding FCRN.
Using PH for decoding achieved a similar effect.
However, when we used a much larger corpus, SLD (about 56 million characters), for decoding, performance significantly improved. A larger and richer corpus made the language model more general and objective, and most importantly, helped to overcome the curse of dimensionality problem\cite{bengio2006neural}.

The RCNN architecture proposed by Shi et al.\cite{shi2015end} is the special case of our FCRN with the path signature truncated at level zero (i.e., Sig0).
As presented in Table~\ref{TableFinalResult}, FCRN outperformed CRNN in both D-Casia and D-Com, which suggested that FCRN captured more essential online information from the pen-tip trajectory and was a better choice for the HCTR problem.
We also observed that our FCRN with naive decoding already achieved comparable results with those of Zhou et al.\cite{zhou2013handwritten}\cite{zhou2014minimum} on dataset D-Casia.
Furthermore, when decoded with the trigram language model based on the SLD corpus, our system outperformed the other methods, with a CR of 96.40\% and an AR of 95.34\% on dataset D-Casia.
On dataset D-Com, although the result can not be strictly compared because of the removal of outlier characters, it may be safe to say that our system achieved state-of-the-art performance.

\section{Conclusion}
This paper presented a novel method of fully convolutional recurrent network (FCRN) for handwritten Chinese text recognition.
The proposed FCRN is an end-to-end architecture that directly used online text data during the training process to solve the HCTR problem, completely avoiding the difficulty of segmentation.
In the experiments, we discovered that the path signature truncated at level two could perfectly capture the pen-tip trajectory of the online text data without significantly increasing the computation during the training process.
At the post-processing stage, we present a refined beam search method that effectively integrated explicit language model to perform decoding and significantly improve the recognition result.
On the test set of CASIA-OLHWDB for online handwritten text recognition, our system outperformed all other methods.
On the test set of ICDAR 2013 Chinese handwriting recognition competition, our system achieved state-of-the-art performance.

%the fully convolutional recurrent network for handwritten Chinese text recognition and the region beam search for incorporating explicit language model to decode recognition result%.
%Our paper demonstrates a novel architecture that directly utilizes online text data in training process to solve the HCTR problem, completely avoiding the difficulty of segmentation.
%Our proposed region beam search method can effectively integrate explicit language model to do decoding and improve recognition result.

%Further work can focus on the improvement of the test set of ICDAR 2011 Chinese handwriting recognition competition and generalization to future work.

In the experiment, our system performed much better on dataset D-Casia than D-Com because of the unbalanced per-character sample distribution on the datasets. Our future work will focus on enriching the training dataset to improve the performance on dataset D-Com using some data augmentation approaches such as sample synthesis or sample distortion.

%This paper present a novel framework for handwritten Chinese text recognition, which consist of two components, including the fully convolutional recurrent network and the region beam search method for language modeling.
%%the fully convolutional recurrent network for handwritten Chinese text recognition and the region beam search for incorporating explicit language model to decode recognition result%.
%Our proposed fully convolutional recurrent network demonstrate a novel architecture that directly utilize online text data in training process to solve the HCTR Problem, completely avoiding the difficulty of segmentation.
%Our proposed region beam search method can effectively integrate explicit language model to decode recognition result.
%In the experiments, we discover that path-signature truncated at level 2 can perfectly capture the pen-tip trajectory of the online text data while not adding too much burden in the training process.
%As is presented in the experiments, Our system outperform all the other methods on the test set of CASIA-OLHWDB for online handwritten text recognition. On the test set of ICDAR 2011 Chinese handwriting recognition competition, our system reach state-of-the-art performance.

\section*{Acknowledgment}
%\footnotesize
This research is supported in part by NSFC (Grant No.: 61472144), GDSTP (Grant No.: 2013B010202004, 2015B010131004) , GDUPS (2011). Research Fund for the Doctoral Program of Higher Education of China (Grant No.: 20120172110023).
%\footnotesize

\bibliographystyle{IEEEtran}
\bibliography{refs}

\end{document}